\documentclass[11pt,a4paper,dvipsnames]{article}
\usepackage[hyperref]{acl2019}
\usepackage{times}
\usepackage{latexsym}
\usepackage{amsmath}
\usepackage{ amssymb }
\usepackage{graphicx}
\usepackage{subcaption}
\usepackage[utf8]{inputenc}
\usepackage{xcolor}
\usepackage{multirow}
\usepackage{array}
\usepackage{booktabs}

\usepackage{url}

\aclfinalcopy 
\def\aclpaperid{678} 

\newcommand\fone{F\textsubscript{1} }

\title{Multi-Hop Paragraph Retrieval for Open-Domain Question Answering}

  \author{Yair Feldman \and Ran El-Yaniv\\
  Department of Computer Science \\
  Technion – Israel Institute of Technology \\
  Haifa, Israel \\
  \texttt{\{yairf11, rani\}@cs.technion.ac.il}}

\date{}

\begin{document}
\maketitle
\begin{abstract}
  This paper is concerned with the task of multi-hop open-domain Question Answering (QA). This task is particularly challenging since it requires the simultaneous performance of textual reasoning and efficient searching. We present a method for retrieving multiple supporting paragraphs, nested amidst a large knowledge base, which contain the necessary evidence to answer a given question. Our method iteratively retrieves supporting paragraphs by forming a joint vector representation of both a question and a paragraph. The retrieval is performed by considering contextualized sentence-level representations of the paragraphs in the knowledge source. Our method achieves state-of-the-art performance over two well-known datasets, SQuAD-Open and HotpotQA, which serve as our single- and multi-hop open-domain QA benchmarks, respectively.
  \footnote{Code is available at https://github.com/yairf11/MUPPET}
\end{abstract}

\section{Introduction}

Textual Question Answering (QA) is the task of answering natural language questions given a set of contexts from which the answers to these questions can be inferred. This task, which falls under the domain of natural language understanding, has been attracting massive interest due to extremely promising results that were achieved using deep learning techniques. These results were made possible by the recent creation of a variety of large-scale QA datasets, such as TriviaQA \citep{JoshiCWZ17triviaqa} and SQuAD \citep{RajpurkarZLL16squad}. The latest state-of-the-art methods are even capable of outperforming humans on certain tasks \citep{bert}\footnote{https://rajpurkar.github.io/SQuAD-explorer/}. \par
The basic and arguably the most popular task of QA is often referred to as Reading Comprehension (RC), in which each question is paired with a relatively small number of paragraphs (or documents) from which the answer can potentially be inferred. The objective in RC is to extract the correct answer from the given contexts or, in some cases, deem the question unanswerable \citep{squad2.0}. Most large-scale RC datasets, however,  are built in such a way that the answer can be inferred using a single paragraph or document. This kind of reasoning is termed \emph{single-hop reasoning}, since it requires reasoning over a single piece of evidence. A more challenging task, called \emph{multi-hop reasoning}, is one that requires combining evidence from multiple sources \cite{TalmorB18complexwebquestions, WelblSR18wikihop, Yang0ZBCSM18hotpot}. Figure \ref{fig:multi_hop} provides an example of a question requiring multi-hop reasoning. To answer the question, one must first infer from the first context that Alex Ferguson is the manager in question, and only then can the answer to the question be inferred with any confidence  from the second context. \par

\begin{figure}[h]
\center
\fbox{\parbox{0.45\textwidth}{
\begin{small}
\textbf{Question:} The \textcolor{OliveGreen}{football manager who recruited David Beckham} \textcolor{RedViolet}{managed Manchester United during what timeframe}?

\textbf{Context 1:} 
The 1995–96 season was Manchester United's fourth season in the Premier League ... Their triumph was made all the more remarkable by the fact that \textcolor{MidnightBlue}{\textbf{\textit{Alex Ferguson}}} ... \textcolor{OliveGreen}{had drafted in young players like Nicky Butt, \textbf{David Beckham}, Paul Scholes and the Neville brothers, Gary and Phil.}

\textbf{Context 2:} Sir \textcolor{MidnightBlue}{\textbf{\textit{Alexander Chapman Ferguson}}}, CBE (born 31 December 1941) is a Scottish former football manager and player \textcolor{RedViolet}{who managed Manchester United \textbf{from 1986 to 2013}.} He is regarded by many players, managers and analysts to be one of the greatest and most successful managers of all time.

\end{small}
}}
\caption{An example of a question and its answer contexts from the HotpotQA dataset requiring multi-hop reasoning and retrieval. The first reasoning hop is highlighted in \textcolor{OliveGreen}{green}, the second hop in \textcolor{RedViolet}{purple}, and the entity connecting the two is highlighted in \textbf{\textit{\textcolor{MidnightBlue}{blue bold italics}}}. In the first reasoning hop, one has to infer that the manager in question is Alex Ferguson. Without this knowledge, the second context cannot possibly be retrieved with confidence, as the question could refer to any of the club's managers throughout its history. Therefore, an \textit{iterative retrieval} is needed in order to correctly retrieve this context pair.}
\label{fig:multi_hop}
\end{figure}

Another setting for QA is \emph{open-domain QA}, in which questions are given without any accompanying contexts, and one is required to locate the relevant contexts to the questions from a large knowledge source (e.g., Wikipedia), and then extract the correct answer using an RC component. This task has recently been resurged following the work of \citet{ChenFWB17drqa}, who used a TF-IDF based retriever to find potentially relevant documents, followed by a neural RC component that extracted the most probable answer from the retrieved documents. While this methodology performs reasonably well for questions requiring single-hop reasoning, its performance decreases significantly when used for open-domain multi-hop reasoning. \par
We propose a new approach to accomplishing this task, called \emph{iterative multi-hop retrieval}, in which one iteratively retrieves the necessary evidence to answer a question. We believe this iterative framework is essential for answering multi-hop questions, due to the nature of their reasoning requirements. \par
Our main contributions are the following:
\begin{itemize}
    \item We propose a novel multi-hop retrieval approach, which we believe is imperative for truly solving the open-domain multi-hop QA task.
    \item We show the effectiveness of our approach, which achieves state-of-the-art results in both single- and multi-hop open-domain QA benchmarks.
    \item We also propose using sentence-level representations for retrieval, and show the possible benefits of this approach over paragraph-level representations.
\end{itemize}
While there are several works that discuss solutions for multi-hop reasoning \cite{dhingra2018multiCoref, zhong2018coarsegrain}, to the best of our knowledge, this work is the first to propose a viable solution for open-domain multi-hop QA.

\section{Task Definition}
We define the open-domain QA task by a triplet $(KS, Q, A)$ where $KS=\{P_1, P_2, \ldots, P_{|KS|}\}$ is a background knowledge source and $P_i = (p_1, p_2, \ldots, p_{l_i})$ is a textual paragraph consisting of $l_i$ tokens, $Q=(q_1, q_2, \ldots, q_{m})$ is a textual question consisting of $m$ tokens, and $A=(a_1, a_2, \ldots, a_{n})$ is a textual answer consisting of $n$ tokens, typically a span of tokens $p_{j_1}, \ldots, p_{j_n}$ in some $P_i \in KS$, or optionally a choice from a predefined set of possible answers. The objective of this task is to find the answer $A$ to the question $Q$ using the background knowledge source $KS$. Formally speaking, our task is to learn a function $\phi$ such that $A = \phi (Q, KS)$.

\paragraph{Single-Hop Retrieval}
In the classic and most simple form of QA, questions are formulated in such a way that the \textit{evidence} required to answer them may be contained in a single paragraph, or even in a single sentence. Thus, in the open-domain setting, it might be sufficient to retrieve a single relevant paragraph $P_i \in KS$ using the information present in the given question $Q$, and have a reading comprehension model extract the answer $A$ from $P_i$. We call this task variation \textit{single-hop retrieval}.

\paragraph{Multi-Hop Retrieval}
In contrast to the single-hop case, there are types of questions whose answers can only be inferred by using at least two different paragraphs. The ability to reason with information taken from more than one paragraph is known in the literature as \textit{multi-hop reasoning} \citep{WelblSR18wikihop}. 
In multi-hop reasoning, not only might the evidence be spread across multiple paragraphs, but it is often necessary to first read a subset of these paragraphs in order to extract the useful information from the other paragraphs, which might otherwise be understood as not completely relevant to the question. This situation becomes even more difficult in the open-domain setting, where one must first find an initial evidence paragraph in order to be able to retrieve the rest. This is demonstrated in Figure \ref{fig:multi_hop}, where one can observe that the second context alone may appear to be irrelevant to the question at hand and the information in the first context is necessary to retrieve the second part of the evidence correctly. 

We extend the multi-hop reasoning ability to the open-domain setting, referring to it as \textit{multi-hop retrieval}, in which the evidence paragraphs are retrieved in an iterative fashion. We focus on this task and limit ourselves to the case where two iterations of retrieval are necessary and sufficient.

\begin{figure*}[!htb]
\centering
\includegraphics[width=0.9\linewidth]{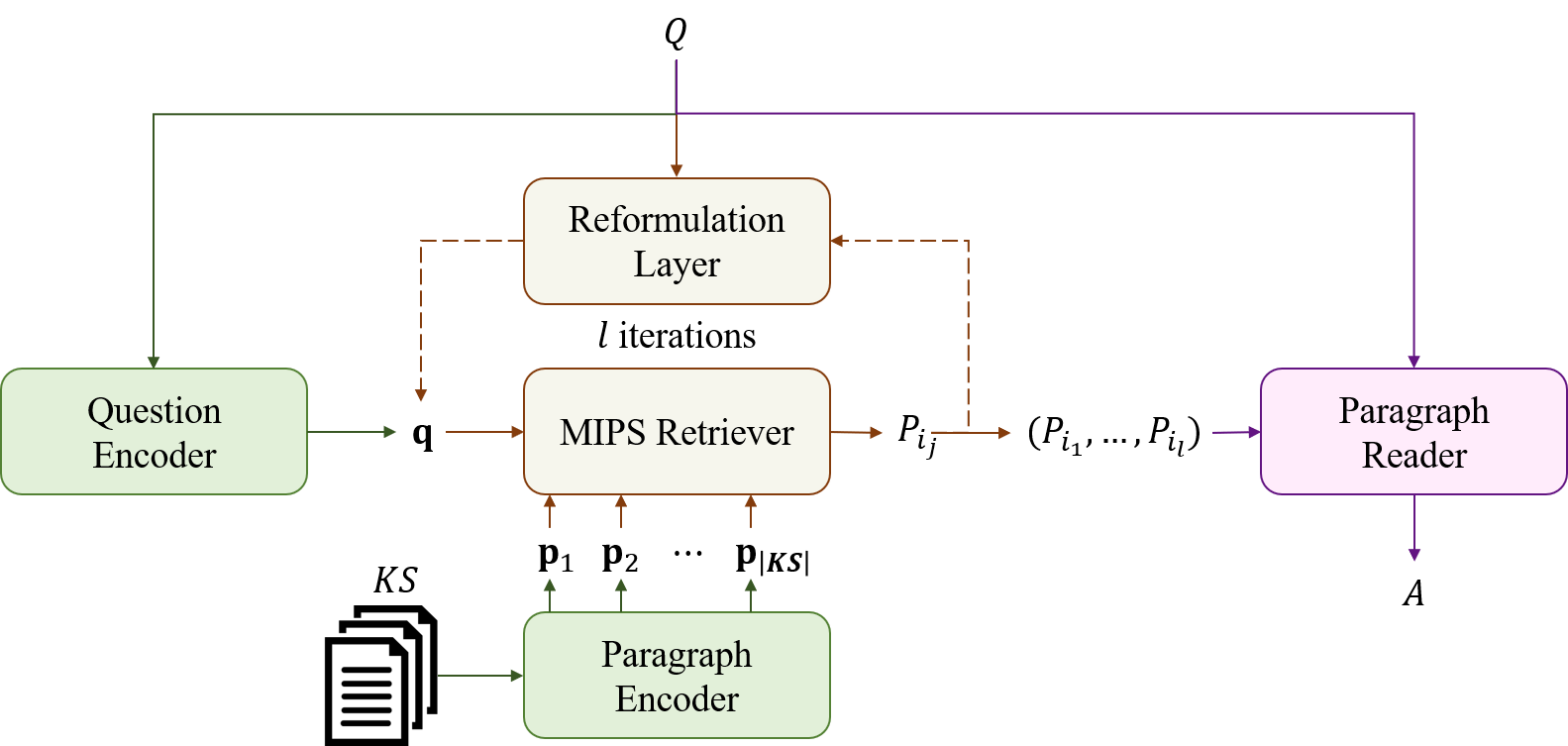}

\caption{A high-level overview of our solution, MUPPET.}
\label{fig:pipeline}
\end{figure*}

\section{Methodology}
Our solution, which we call MUPPET (\textbf{mu}lti-ho\textbf{p} \textbf{p}aragraph r\textbf{et}rieval), relies on the following \emph{basic scheme} consisting of two main components: (a) a paragraph and question encoder, and (b) a paragraph reader. The encoder is trained to encode paragraphs into $d$-dimensional vectors, and to encode questions into \emph{search vectors} in the same vector space. Then, a \emph{maximum inner product search} (MIPS) algorithm is applied to find the most similar paragraphs to a given question. Several algorithms exist for fast (and possibly approximate) MIPS, such as the one proposed by \citet{JohnsonDJ17faiss}.  The most similar paragraphs are then passed to the paragraph reader, which, in turn, extracts the most probable answer to the question. \par
It is critical that the paragraph encodings do not depend on the questions. This enables storing precomputed paragraph encodings and executing efficient MIPS when given a new search vector. Without this property, any new question would require the processing of the complete knowledge source (or a significant part of it). \par

To support multi-hop retrieval, we propose the following extension to the basic scheme. Given a question $Q$, we first obtain its encoding $\mathbf{q} \in \mathbb{R}^d$ using the encoder. Then, we transform it into a search vector $\mathbf{q}^s \in \mathbb{R}^d$, which is used to retrieve the top-$k$ relevant paragraphs $\{P^Q_1, P^Q_2, \ldots, P^Q_k\} \subset KS$ using MIPS. In each subsequent retrieval iteration, we use the paragraphs retrieved in its previous iteration to reformulate the search vector. This produces $k$ new search vectors, $\{\tilde{\mathbf{q}}^s_1, \tilde{\mathbf{q}}^s_2, \ldots, \tilde{\mathbf{q}}^s_k\}$, where $\tilde{\mathbf{q}}^s_i \in \mathbb{R}^d$, which are used in the same manner as in the first iteration to retrieve the next top-$k$ paragraphs, again using MIPS.
This method can be seen as performing a beam search of width $k$ in the encoded paragraphs' space. A high-level view of the described solution is given in Figure \ref{fig:pipeline}. \par

\begin{figure}[h]
\centering
\begin{subfigure}{0.4\linewidth}
\centering
\includegraphics[height=5.5cm]{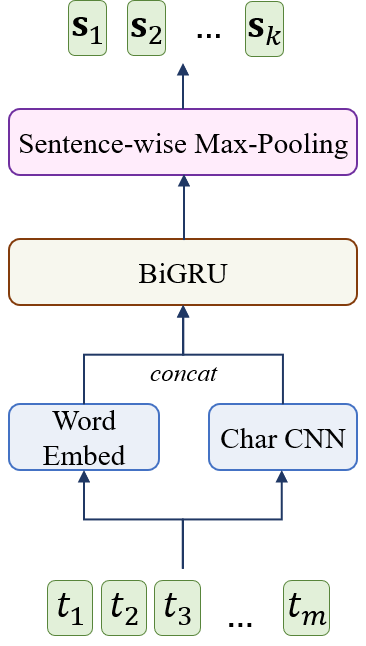}
\caption{Sentence Encoder}
\label{fig:encoder}
\end{subfigure}
\hspace{0.1cm}
\begin{subfigure}{0.5\linewidth}
\centering
\includegraphics[height=5.5cm]{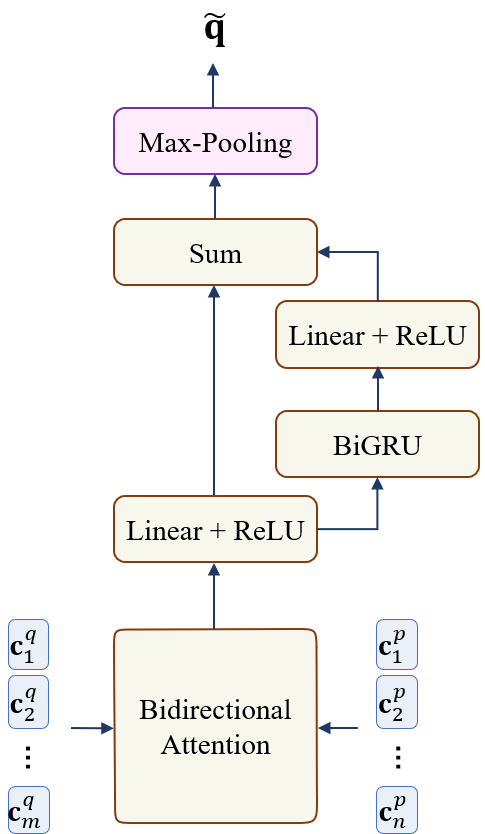}
\caption{Reformulation Component}
\label{fig:reform}
\end{subfigure}
\caption{Architecture of the main components of our paragraph and question encoder. (a) Our sentence encoder architecture. The model receives a series of tokens as input and produces a sequence of sentence representations. (b) Our reformulation component architecture. This layer receives contextualized representations of a question and a paragraph, and produces a reformulated representation of the question.}
\label{fig:architectures}
\end{figure}

\subsection{Paragraph and Question Encoder}
\label{sec:encoder}
We define $f$, our encoder model, in the following way. 
Given a paragraph $P$ consisting of $k$ sentences $(s_1, s_2, \ldots, s_k)$ and $m$ tokens $(t_1, t_2, \ldots, t_m)$, such that $s_i = (t_{i_1}, t_{i_2}, \ldots, t_{i_l})$, where $l$ is the length of the sentence, our encoder generates $k$ respective $d$-dimensional encodings $(\mathbf{s}_1, \mathbf{s}_2, \ldots, \mathbf{s}_k) = f(P)$, one for each sentence. This is in contrast to previous work in paragraph retrieval in which only a single fixed-size representation is used for each paragraph \citep{LeeYKKK18answerrecall, das2018multistep}. The encodings are created by passing $(t_1, t_2, \ldots, t_m)$ through the following layers.

\subparagraph{Word Embedding}
We use the same embedding layer as the one suggested by \citet{GardnerC18snorm}. Each token $t$ is embedded into a vector $\mathbf{t}$ using both character-level and word-level information. The word-level embedding $\mathbf{t}^w$ is obtained via pretrained word embeddings. The character-level embedding of a token $t$ with $l_t$ characters $(t^c_1, t^c_2, \ldots, t^c_{l_t})$ is obtained in the following manner: each character $t^c_i$ is embedded into a fixed-size vector $\mathbf{t}^c_i$. We then pass each token's character embeddings through a one-dimensional convolutional neural network, followed by max-pooling over the filter dimension. This produces a fixed-size character-level representation for each token, $\mathbf{t}^c = \max \big(\text{CNN}(\mathbf{t}^c_1, \textbf{t}^c_2, \ldots, \textbf{t}^c_{l_t})\big)$.  Finally, we concatenate the word-level and character-level embeddings to form the final word representation, $\mathbf{t} = [\mathbf{t}^w; \mathbf{t}^c]$.

\subparagraph{Recurrent Layer}
After obtaining the word representations, we use a bidirectional GRU \citep{ChoMGBBSB14gru} to process the paragraph and obtain the contextualized word representations, $$(\mathbf{c}_1, \mathbf{c}_2, \ldots, \mathbf{c}_m) = \text{BiGRU}(\mathbf{t}_1, \mathbf{t}_2, \ldots, \mathbf{t}_m).$$

\subparagraph{Sentence-wise max-pooling}
Finally, we chunk the contextualized representations of the paragraph tokens into their corresponding sentence groups, and apply max-pooling over the time dimension of each sentence group to obtain the parargaph's $d$-dimensional sentence representations, $\mathbf{s}_i = \max (\mathbf{c}_{i_1}, \mathbf{c}_{i_2}, \ldots, \mathbf{c}_{i_l})$.
A high-level outline of the sentence encoder is shown is Figure \ref{fig:encoder}, where we can see a series of $m$ tokens being passed through the aforementioned layers, producing $k$ sentence representations.
\par
The encoding $\mathbf{q}$ of a question $Q$ is computed similarly, such that $\mathbf{q} = f(Q)$. Note that we produce a single vector for any given question, thus the max-pooling operation is applied over all question words at once, disregarding sentence information.

\begin{figure}[h]
\center
\fbox{\parbox{0.45\textwidth}{
\begin{small}
\textbf{Context:} 
\textcolor{RedViolet}{One of the most famous people born in Warsaw was Maria Skłodowska-Curie, who achieved international recognition for her research on radioactivity and was the first female recipient of the Nobel Prize.} Famous musicians include Władysław Szpilman and Frédéric Chopin. \textcolor{OliveGreen}{Though Chopin was born in the village of Żelazowa Wola, about 60 km (37 mi) from Warsaw, he moved to the city with his family when he was seven months old.} Casimir Pulaski, a Polish general and hero of the American Revolutionary War, was born here in 1745.

\textbf{Question 1:} \textcolor{RedViolet}{What was Maria Curie the first female recipient of?}

\textbf{Question 2:} \textcolor{OliveGreen}{How old was Chopin when he moved to Warsaw with his family?}

\end{small}
}}
\caption{An example from the SQuAD dataset of a paragraph that acts as the context for two different questions. Question 1 and its evidence (highlighted in \textcolor{RedViolet}{purple}) have little relation to question 2 and its evidence (highlighted in \textcolor{OliveGreen}{green}). This motivates our method of storing sentence-wise encodings instead of a single representation for an entire paragraph.}
\label{fig:sentence_level}
\end{figure}

\subparagraph{Reformulation Component}
The reformulation component receives a paragraph $P$ and a question $Q$, and produces a single vector $\tilde{\mathbf{q}}$. First, contextualized word representations are obtained using the same embedding and recurrent layers used for the initial encoding, $(\mathbf{c}^q_1, \mathbf{c}^q_2, \ldots, \mathbf{c}^q_{n_q})$ for $Q$ and $(\mathbf{c}^p_1, \mathbf{c}^p_2, \ldots, \mathbf{c}^p_{n_p})$ for $P$.
We then pass the contextualized representations through a \textbf{bidirectional attention} layer, which we adopt from \citet{GardnerC18snorm}.
The attention between question word $i$ and paragraph word $j$ is computed as:
$$a_{ij} = \textbf{w}_1^a \cdot \textbf{c}^q_i + \textbf{w}_2^a \cdot \textbf{c}^p_j + \textbf{w}_3^a \cdot (\textbf{c}^q_i \odot \textbf{c}^p_j),$$
where $\textbf{w}_1^a, \textbf{w}_2^a, \textbf{w}_3^a \in \mathbb{R}^d$ are learned vectors. For each question word, we compute the vector $\textbf{a}_i$:
\begin{equation*}
\alpha_{ij}=\frac{e^{a_{ij}}}{\sum^{n_p}_{j=1} e^{a_{ij}}}
,\quad 
\textbf{a}_i = \sum^{n_p}_{j=1} \alpha_{ij}\textbf{c}^p_j.
\end{equation*}
A paragraph-to-question vector $\textbf{a}^p$ is computed as follows:
\begin{equation*}
m_i = \max_{1\leq j \leq n_p} a_{ij}
,\quad 
\beta_i = \frac{e^{m_i}}{\sum^{n_q}_{i=1} e^{m_i}}
\end{equation*}
$$\textbf{a}^p = \sum^{n_q}_{i=1} \beta_i\textbf{c}^q_i.$$
We concatenate $\textbf{c}^q_i, \textbf{a}_i, \textbf{c}^q_i \odot \textbf{a}_i$ and $\textbf{a}^p \odot \textbf{a}_i$ and pass the result through a linear layer with ReLU activations to compute the final bidirectional attention vectors. We also use a residual connection where we process these representations with a bidirectional GRU and another linear layer with ReLU activations. Finally, we sum the outputs of the two linear layers. As before, we derive the $d$-dimensional reformulated question representation $\tilde{\textbf{q}}$ using a max-pooling layer on the outputs of the residual layer. A high-level outline of the reformulation layer is given in Figure \ref{fig:reform}, where $m$ contextualized token representations of the question and $n$ contextualized token representations of the paragraph are passed through the component's layers to produce the reformulated question representation, $\tilde{\textbf{q}}$.

\subparagraph{Relevance Scores}
Given the sentence representations $(\mathbf{s}_1, \mathbf{s}_2, \ldots, \mathbf{s}_k)$ of a paragraph $P$, and the question encoding $\mathbf{q}$ for $Q$, the relevance score of $P$ with respect to a question $Q$ is calculated in the following way:
$$ rel(Q, P) = \max_{i=1,\ldots,k}{\sigma\Bigg(\begin{bmatrix}\mathbf{s}_i \\ \mathbf{s}_i \odot \mathbf{q} \\ \mathbf{s}_i \cdot \mathbf{q} \\ \mathbf{q}\end{bmatrix} \cdot \begin{bmatrix}\mathbf{w}_1 \\ \mathbf{w}_2 \\ w_3 \\ \mathbf{w}_4 \end{bmatrix} + b\Bigg),} $$
where $\mathbf{w}_1, \mathbf{w}_2, \mathbf{w}_4 \in \mathbb{R}^d$ and $w_3, b \in \mathbb{R}$ are  learned parameters. \par
A similar max-pooling encoding approach, along with the scoring layer's structure, were proposed by \citet{ConneauKSBB17snli} who showed their efficacy on various sentence-level tasks.
We find this sentence-wise formulation to be beneficial because it suffices for one sentence in a paragraph to be relevant to a question for the whole paragraph to be considered as relevant. This allows more fine-grained representations for paragraphs and more accurate retrieval. An example of the benefits of using this kind of sentence-level model is given in Figure \ref{fig:sentence_level}, where we see two questions answered by two different sentences. Our model allows each question to be similar only to parts of the paragraph, and not necessarily to all of it. \par
\subparagraph{Search Vector Derivation}
Recall that our retrieval algorithm is based on executing a MIPS in the paragraph encoding space. To derive such a search vector from the question encoding $\mathbf{q}$, we observe that:
$$ rel(Q, P) \propto \max_{i=1,\ldots,k}{\mathbf{s}_i^\top (\mathbf{w}_1 + \mathbf{w}_2 \odot \mathbf{q} + w_3 \cdot \mathbf{q}). } $$
Therefore, the final search vector of a question $Q$ is $\mathbf{q}_s = \mathbf{w}_1 + \mathbf{w}_2 \odot \mathbf{q} + w_3 \cdot \mathbf{q}$. The same equations apply when predicting the relevance score for the second retrieval iteration, in which case $\mathbf{q}$ is swapped with $\tilde{\mathbf{q}}$. 

\paragraph{Training and Loss Functions}
Each training sample consists of a question and two paragraphs, $(Q, P^1, P^2)$, where $P^1$ corresponds to a paragraph retrieved in the first iteration, and $P^2$ corresponds to a paragraph retrieved in the second iteration using the reformulated vector $\tilde{\mathbf{q}}$. $P^1$ is considered relevant if it constitutes one of the necessary evidence paragraphs to answer the question. $P^2$ is considered relevant only if $P^1$ and $P^2$ together constitute the complete set of evidence paragraphs needed to answer the question. Both iterations have the same form of loss functions, and the model is trained by optimizing the sum of the iterations' losses. \par
Our training objective for each iteration is composed of two components: a binary cross-entropy loss function and a ranking loss function.
The cross-entropy loss is defined as follows:
\begin{equation*}
\begin{split}
\mathcal{L}_{CE} = -\frac{1}{N}&\sum_{i=1}^{N} y_i \log{\big(rel(Q_i, P_i)\big)} \\ 
& + (1-y_i) \log{\big(1-rel(Q_i, P_i)\big),}
\end{split}
\end{equation*}
where $y_i \in \{0,1\}$ is a binary label indicating the true relevance of $P_i$ to $Q_i$ in the iteration in which $rel(Q_i, P_i)$ is calculated, and $N$ is the number of samples in the current batch. \par
The ranking loss is computed in the following manner. First, for each question $Q_i$ in a given batch, we find the mean of the scores given to positive and negative paragraphs for each question, $ q^{pos}_i = \frac{1}{M_1}\sum_{j=1}^{M_1} rel(Q_i, P_j)$ and $ q^{neg}_i = \frac{1}{M_2}\sum_{j=1}^{M_2} rel(Q_i, P_j)$, where $M_1$ and $M_2$ are the number of positive and negative samples for $Q_i$, respectively. We then define the margin ranking loss \citep{socher2013marginLoss} as
\begin{equation}
\label{eq:rank_obj}
    \mathcal{L}_{R} = \frac{1}{M}\sum_{i=1}^{M} \max (0, \gamma - q^{pos}_i + q^{neg}_i),
\end{equation}
where $M$ is the number of distinct questions in the current batch, and $\gamma$ is a hyperparameter. The final objective is the sum of the two losses:
\begin{equation}
\label{eq:sum_obj}
    \mathcal{L} = \mathcal{L}_{CE} + \lambda\mathcal{L}_{R},
\end{equation}
where $\lambda$ is a hyperparameter. \par
We note that we found it slightly beneficial to incorporate pretrained ELMo \citep{Peters2018elmo} embeddings in our model.
For more detailed information of the implementation details and training process, please refer to Appendix \ref{sec:implementation}.

\subsection{Paragraph Reader} \label{reader}
The paragraph reader receives as input a question $Q$ and a paragraph $P$ and extracts the most probable answer span to $Q$ from $P$.
We use the S-norm model proposed by \citet{GardnerC18snorm}. A detailed description of the model is given in Appendix \ref{sec:sup_reader}.

\paragraph{Training}
An input sample for the paragraph reader consists of a question and a single context $(Q, P)$. 
We optimize the same negative log-likelihood function used in the S-norm model for the span start boundaries:
\begin{equation*}
\begin{split}
\mathcal{L}_{start} = -\log\Bigg(\frac{\sum_{j\in P^Q} \sum_{k\in A_j} e^{s_{kj}}}{\sum_{j\in P^Q} \sum_{i=1}^{n_j} e^{s_{ij}}}\Bigg),
\end{split}
\end{equation*}
where $P^Q$ is the set of paragraphs paired with the same question $Q$, $A_j$ is the set of tokens that start an answer span in the $j$-th paragraph, and $s_{ij}$ is the score given to the $i$-th token in the $j$-th paragraph. The same formulation is used for the span end boundaries, so that the final objective function is the sum of the two: $\mathcal{L}_{span} = \mathcal{L}_{start} + \mathcal{L}_{end}$. \par

\section{Experiments and Results}
We test our approach on two datasets, and measure end-to-end QA performance using the standard \emph{exact match} (EM) and \fone metrics, as well as the metrics proposed by \citet{Yang0ZBCSM18hotpot} for the HotpotQA dataset (see Appendix \ref{sec:hotpot_extension}).
\subsection{Datasets}
\paragraph{HotpotQA} \citet{Yang0ZBCSM18hotpot} introduced a dataset of Wikipedia-based questions, which require reasoning over multiple paragraphs to find the correct answer. The dataset also includes hard supervision on sentence-level supporting facts, which encourages the model to give explainable answer predictions. Two benchmark settings are available for this dataset: (1) a \emph{distractor} setting, in which the reader is given a question as well as a set of paragraphs that includes both the supporting facts and irrelevant paragraphs; (2) a \emph{full wiki} setting, which is an open-domain version of the dataset. We use this dataset as our benchmark for the \textit{multi-hop retrieval} setting. Several extensions must be added to the reader from Section \ref{reader} in order for it to be suitable for the HotpotQA dataset. A detailed description of our proposed extensions is given in Appendix \ref{sec:hotpot_extension}.

\paragraph{SQuAD-Open} \citet{ChenFWB17drqa} decoupled the questions from their corresponding contexts in the original SQuAD dataset \citep{RajpurkarZLL16squad}, and formed an open-domain version of the dataset by defining an entire Wikipedia dump to be the background knowledge source from which the answer to the question should be extracted. We use this dataset to test the effectiveness of our method in a classic \textit{single-hop retrieval} setting.

\subsection{Experimental Setup}
\subparagraph{Search Hyperparameters}
For our experiments in the multi-hop setting, we used a width of 8 in the first retrieval iteration. In all our experiments, unless stated otherwise, the reader is fed the top 45 paragraphs through which it reasons independently and finds the most probable answers. In addition, we found it beneficial to limit the search space of our MIPS retriever to a subset of the knowledge source, which is determined by a TF-IDF heuristic retriever. We define $n_i$ to be the size of the search space for retrieval iteration $i$. As we will see, there is a trade-off for choosing various values of $n_i$. A large value of $n_i$ offers the possibility of higher recall, whereas a small value of $n_i$ introduces less noise in the form of irrelevant paragraphs. \par
\subparagraph{Knowledege Sources}
For HotpotQA, our knowledge source is the same Wikipedia version used by \citet{Yang0ZBCSM18hotpot}\footnote{It has recently come to our attention that during our work, some details of the Wikipedia version have changed. Due to time limitations, we use the initial version description.}. This version is a set of all of the first paragraphs in the entire Wikipedia. For SQuAD-Open, we use the same Wikipedia dump used by \citet{ChenFWB17drqa}. For both knowledge sources, the TF-IDF based retriever we use for search space reduction is the one proposed by \citet{ChenFWB17drqa}, which uses bigram hashing and TF-IDF matching. We note that in the HotpotQA Wikipedia version each document is a single paragraph, while in SQuAD-Open, the full Wikipedia documents are used. 
\par 

\begin{table*}[!ht]
    \small
    \centering
    \begin{tabular}{llcccccc}
        \toprule
        \multirow{2}{*}{Setting} & \multirow{2}{*}{Method} & \multicolumn{2}{c}{Answer} & \multicolumn{2}{c}{Sup Fact} & \multicolumn{2}{c}{Joint} \\
        \cmidrule{3-8}
        & & EM & \fone & EM & \fone & EM & \fone \\
        \midrule
        \multirow{2}{*}{distractor} 
        & Baseline \citep{Yang0ZBCSM18hotpot} & 44.44 & 58.28 & 21.95 & 66.66 & 11.56 & 40.86 \\
        \cmidrule{2-8}
        & Our Reader & 51.56 & 65.32 & 44.54 & 75.27 & 28.68 & 54.08 \\
        \midrule
        \multirow{4}{*}{full wiki} 
        & Baseline \citep{Yang0ZBCSM18hotpot} & 24.68 & 34.36 & \phantom{0}5.28 & 40.98 & \phantom{0}2.54 & 17.73 \\
        \cmidrule{2-8}
        & TF-IDF + Reader & 27.55 & 36.58 & 10.75 & 42.45 & \phantom{0}7.00 & 21.47 \\
        & MUPPET (sentence-level) & 30.20 & 39.43 & 16.57 & 46.13 & 11.38 & 26.55 \\
        & MUPPET (paragraph-level) & \textbf{31.07} & \textbf{40.42} & \textbf{17.00} & \textbf{47.71} & \textbf{11.76} & \textbf{27.62} \\
        \bottomrule
    \end{tabular}
    \caption{Primary results for HotpotQA (dev set). At the top of the table, we compare our Paragraph Reader to the baseline model of \citet{Yang0ZBCSM18hotpot} (as of writing this paper, no other published results are available other than the baseline results). At the bottom, we compare the end-to-end performance on the full wiki setting. TF-IDF + Reader refers to using the TF-IDF based retriever without our MIPS retriever. MUPPET (sentence-level) refers to our approach with sentence-level representations, and MUPPET (paragraph-level) refers to our approach with paragraph-level representations. For both sentence- and paragraph-level results, we set $n_1 = 32$ and $n_2 = 512$. } 
    \label{tab:primary_hotpot}
\end{table*}

\begin{table}[h]
    \small
    \centering
    \begin{tabular}{lcc}
        \toprule
        Method & EM & \fone \\
        \midrule
        DrQA \citep{ChenFWB17drqa} & 28.4 & - \\
        DrQA \citep{ChenFWB17drqa} (multitask) & 29.8 & - \\
        R\textsuperscript{3} \citep{WangYGWKZCTZJ18r3} & 29.1 & 37.5 \\
        DS-QA \citep{SunLLJ18denoise} & 28.7 & 36.6 \\
        Par. Ranker + Full Agg. \citep{LeeYKKK18answerrecall} & 30.2 & - \\
        Minimal \citep{SocherZXM18minimal} & 34.7 & 42.6 \\
        Multi-step \citep{das2018multistep} & 31.9 & 39.2 \\
        BERTserini
        \citep{yang2019bert_serini} & 38.6 & 46.1 \\
        \cmidrule{1-3}
        TF-IDF + Reader & 34.6 & 41.6  \\
        MUPPET (sentence-level) & \textbf{39.3} & \textbf{46.2} \\
        MUPPET (paragraph-level) & 35.6 & 42.5 \\
        \bottomrule
    \end{tabular}
    \caption{Primary results for SQuAD-Open. } 
    \label{tab:primary_squad}
\end{table}

\subsection{Results}
\paragraph{Primary Results}
Tables \ref{tab:primary_hotpot} and \ref{tab:primary_squad} show our main results on the HotpotQA and SQuAD-Open datasets, respectively. In the HotpotQA distractor setting, our paragraph reader greatly improves the results of the baseline reader, increasing the joint EM and \fone scores by $17.12$ (148\%) and $13.22$ (32\%) points, respectively. 
In the full wiki setting, we compare three methods of retrieval: (1) TF-IDF, in which only the TF-IDF heuristic is used. The reader is fed all possible paragraph pairs from the top-$10$ paragraphs. (2) Sentence-level, in which we use MUPPET with sentence-level encodings. (3) Paragraph-level, in which we use MUPPET with paragraph-level encodings (no sentence information).
We can see that both methods significantly outperform the na\"ive TF-IDF retriever, indicating the efficacy of our approach. As of writing this paper, we are placed second in the HotpotQA full wiki setting (test set) leaderboard\footnote{March 5, 2019. Leaderboard available at https://hotpotqa.github.io/}. For SQuAD-Open, our sentence-level method established state-of-the-art results, improving the current non-BERT \citep{bert} state-of-the-art by $4.6$ (13\%) and $3.6$ (8\%) EM and \fone points, respectively. This shows that our encoder can be useful not only for multi-hop questions, but also for single-hop questions.

\paragraph{Retrieval Recall Analysis}

\begin{figure*}[!ht]
\centering
\begin{subfigure}{0.3\linewidth}
\centering
\includegraphics[width=\linewidth]{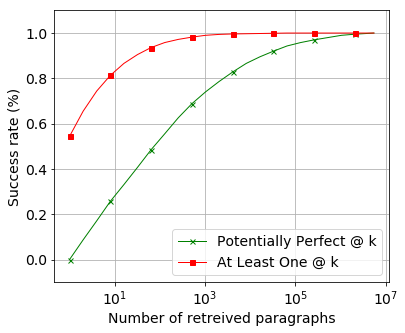}
\caption{TF-IDF retrieval results}
\label{fig:tfidf}
\end{subfigure}
\begin{subfigure}{0.3\linewidth}
\centering
\includegraphics[width=\linewidth]{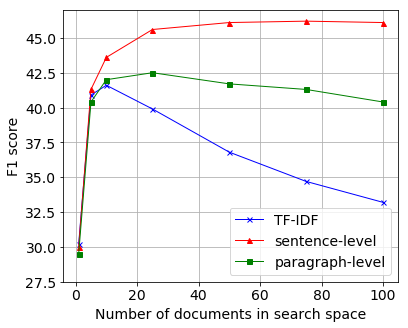}
\caption{SQuAD-Open}
\label{fig:squad_search_space}
\end{subfigure}
\begin{subfigure}{0.3\linewidth}
\centering
\includegraphics[width=\linewidth]{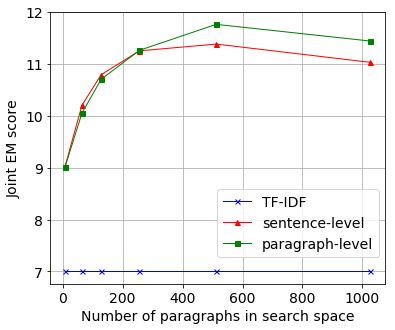}
\caption{HotpotQA}
\label{fig:hotpot_search_space}
\end{subfigure}
\caption{Various results based on the TF-IDF retriever. (a) Retrieval results of the TF-IDF hueristic retriever on HotpotQA. \textit{At Least One @ k} is the number of questions for which at least one of the paragraphs containing the supporting facts is retrieved in the top-$k$ paragraphs. \textit{Potentially Perfect @ k} is the number of questions for which both of the paragraphs containing the supporting facts are retrieved in the top-$k$ paragraphs.
(b) and (c)
Performance analysis on the SQuAD-Open and HotpotQA datasets, respectively, as more documents/paragraphs are retrieved by the TF-IDF heuristic retriever. Note that for SQuAD-Open each document contains several paragraphs, and the reader is fed the top-$k$ TF-IDF ranked paragraphs from within the documents in the search space.}
\label{fig:all_plots}
\end{figure*}

We analyze the performance of the TF-IDF retriever for HotpotQA in Figure \ref{fig:tfidf}. We can see that the retriever succeeds in retrieving at least one of the gold paragraphs for each question (above $90\%$ with the top-$32$ paragraphs), but fails at retrieving both gold paragraphs. This demonstrates the necessity of an efficient multi-hop retrieval approach to aid or replace classic information retrieval methods. 

\paragraph{Effect of Narrowing the Search Space}

In Figures \ref{fig:squad_search_space} and \ref{fig:hotpot_search_space}, we show the performance of our method as a function of the size of the search space of the last retrieval iteration. For SQuAD-Open, the TF-IDF retriever initially retrieves a set of documents, which are then split into paragraphs to form the search space. Each search space of top-$k$ paragraphs limits the potential recall of the model to that of the top-$k$ paragraphs retrieved by the TF-IDF retriever. This proves to be suboptimal for very small values of $k$, as the performance of the TF-IDF retriever is not good enough. Our models, however, fail to benefit from increasing the search space indefinitely, hinting that they are not as robust to noise as we would want them to be.

\paragraph{Effectiveness of Sentence-Level Encodings}
Our method proposes using sentence-level encodings for paragraph retrieval. We test the significance of this approach in Figures \ref{fig:squad_search_space} and \ref{fig:hotpot_search_space}. While sentence-level encodings seem to be vital for improving state-of-the-art results on SQuAD-Open, the same cannot be said about HotpotQA. We hypothesize that this is a consequence of the way the datasets were created. In SQuAD, each paragraph serves as the context of several questions, as shown in Figure \ref{fig:sentence_level}. This leads to questions being asked about facts less essential to the gist of the paragraph, and thus they would not be encapsulated in a single paragraph representation. In HotpotQA, however, most of the paragraphs in the training set serve as the context of at most one question.

\section{Related Work}
\citet{ChenFWB17drqa} first introduced the use of neural methods to the task of open-domain QA using a textual knowledge source. They proposed DrQA, a pipeline approach with two components: a TF-IDF based retriever, and a multi-layer neural network that was trained to find an answer span given a question and a paragraph.
In an attempt to improve the retrieval of the TF-IDF based component, many existing works have used Distant Supervision (DS) to further re-rank the retrieved paragraphs \citep{HtutBC18ranking, Yan2018deepCascade}.
\citet{WangYGWKZCTZJ18r3} used reinforcement learning to train a re-ranker and an RC component in an end-to-end manner, and showed its advantage over the use of DS alone. 
\citet{SocherZXM18minimal} trained a sentence selector and demonstrated the effectiveness of reading minimal contexts instead of complete documents. As DS can often lead to wrong labeling, \citet{SunLLJ18denoise} suggested a denoising method for alleviating this problem. While these methods have proved to increase performance in various open-domain QA datasets, their re-ranking approach is limited in the number of paragraphs it can process, as it requires the joint reading of a question with all possible paragraphs. This is in contrast to our approach, in which all paragraph representations are precomputed to allow efficient large-scale retrieval. There are some works that adopted a similar precomputation scheme.
\citet{LeeYKKK18answerrecall} learned an encoding function for questions and paragraphs and ranked paragraphs by their dot-product similarity with the question. Many of their improvements, however, can be attributed to the incorporation of answer aggregation methods as suggested by \citet{wang2018evidence} in their model, which enhanced their results significantly.
\citet{SeoKPFH18piqa} proposed phrase-indexed QA (PI-QA), a new formulation of the QA task that requires the independent encoding of answers and questions. The question encodings are then used to retrieve the correct answers by performing MIPS. This is more of a challenge task rather than a solution for open-domain QA.
A recent work by \citet{das2018multistep} proposed a new framework for open-domain QA that employs a multi-step interaction between a retriever and a reader. This interactive framework is used to refine a question representation in order for the retrieval to be more accurate. Their method is complimentary to ours -- the interactive framework is used to enhance retrieval performance for single-hop questions, and does not handle the multi-hop domain. \par
Another line of work reminiscent of our method is the one of Memory Networks \citep{memory_networks}. Memory Networks consist of an array of cells, each capable of storing a vector, and four modules (input, update, output and response) that allow the manipulation of the memory for the task at hand. Many variations of Memory Networks have been proposed, such as end-to-end Memory Networks \citep{end-to-end-memory-networks}, Key-Value Memory Networks \citep{miller-etal-2016-key-value}, and Hierarchical Memory Networks \citep{chandar2016hierarchical}.

\section{Concluding Remarks}
We present MUPPET, a novel method for multi-hop paragraph retrieval, and show its efficacy in both single- and multi-hop QA datasets. One difficulty in the open-domain multi-hop setting is the lack of supervision, a difficulty that in the single-hop setting is alleviated to some extent by using distant supervision. We hope to tackle this problem in future work to allow learning more than two retrieval iterations. 
An interesting improvement to our approach would be to allow the retriever to automatically determine whether or not more retrieval iterations are needed. A promising direction could be a multi-task approach, in which both single- and multi-hop datasets are learned jointly. We leave this for future work.

\section*{Acknowledgments}
This research was partially supported by the Israel Science Foundation (grant No. 710/18).

\bibliography{acl2019}
\bibliographystyle{acl_natbib}

\newpage

\appendix

\section{Paragraph Reader}
\label{sec:sup_reader}
In this section we describe in detail the reader mentioned in Section \ref{reader}.
The paragraph reader receives as input a question $Q$ and a paragraph $P$ and extracts the most probable answer span to $Q$ from $P$.
We use the shared-norm model presented by \citet{GardnerC18snorm}, which we refer to as S-norm. The model's architecture is quite similar to the one we used for the encoder. First, we process $Q$ and $P$ seperately to obtain their contexualized token representations, in the same manner as used in the encoder. We then pass the contextualized representations through a bidirectional attention layer similar to the one defined in the reformulation layer of the encoder, with the only difference being that the roles of the question and the paragraph are switched. As before, we further pass the bidirectional attention representations through a residual connection, this time using a self-attention layer between the bidirectional GRU and the linear layer. The self-attention mechanism is similar to the bidirectional attention layer, only now it is between the paragraph and itself. Therefore, question-to-parargaph attention is not used, and we set $a_{ij}=- \infty $ if $i=j$. The summed outputs of the residual connection are passed to the prediction layer. The inputs to the prediction layer are passed through a bidirectional GRU followed by a linear layer that predicts the answer span start scores. The hidden layers of that GRU are concatenated with the input and passed through another bidirectional GRU and linear layer to predict the answer span end scores.

\paragraph{Training}
An input sample for the paragraph reader consists of a question and a single context $(Q, P)$. 
We optimize the same negative log-likelihood function used in the S-norm model for the span start boundaries:
\begin{equation*}
\begin{split}
\mathcal{L}_{start} = -\log\Bigg(\frac{\sum_{j\in P^Q} \sum_{k\in A_j} e^{s_{kj}}}{\sum_{j\in P^Q} \sum_{i=1}^{n_j} e^{s_{ij}}}\Bigg),
\end{split}
\end{equation*}
where $P^Q$ is the set of paragraphs paired with the same question $Q$, $A_j$ is the set of tokens that start an answer span in the $j$-th paragraph, and $s_{ij}$ is the score given to the $i$-th token in the $j$-th paragraph. The same formulation is used for the span end boundaries, so that the final objective function is the sum of the two: $\mathcal{L}_{span} = \mathcal{L}_{start} + \mathcal{L}_{end}$.

\section{Paragraph Reader Extension for HotpotQA}
\label{sec:hotpot_extension}
HotpotQA presents the challenge of not only predicting an answer span, but also yes/no answers. This is a combination of span-based questions and multiple-choice questions. In addition, one is also required to provide explainability to the answer predictions by predicting the supporting facts leading to the answer. We extend the paragraph reader from Section \ref{reader} to support these predictions in the following manner.
\subparagraph{Yes/No Prediction}
We argue that one can decide whether the answer to a given question should be span-based or yes/no-based without looking at any context at all. Therefore, we first create a fixed-size vector representing the question using max-pooling over the first bidirectional GRU's states of the question. We pass this representation through a linear layer that predicts whether this is a yes/no-based question or a span-based question. If span-based, we predict the answer span from the context using the original span prediction layer. If yes/no-based, we encode the question-aware context representations to a fixed-size vector by performing max-pooling over the outputs of the residual self-attention layer. As before, we then pass this vector through a linear layer to predict a yes/no answer.
\subparagraph{Supporting Fact Prediction}
As a context's supporting facts for a question are at the sentence-level, we encode the question-aware context representations to fixed-size sentence representations by passing the outputs of the residual self-attention layer through another bidirectional GRU, followed by performing max-pooling over the sentence groups of the GRU's outputs. Each sentence representation is then passed through a multilayer perceptron with a single hidden layer equipped with ReLU activations to predict whether it is indeed a supporting fact or not.

\paragraph{Training}
An input sample for the paragraph reader consists of a question and a single context, $(Q, P)$. Nevertheless, as HotpotQA requires multiple paragraphs to answer a question, we define $P$ to be the concatenation of these paragraphs. \par
Our objective function comprises four loss functions, corresponding to the four possible predictions of our model. For the span-based prediction we use $\mathcal{L}_{span}$, as before. 
We use a similar negative log likelihood loss for the answer type prediction (whether the answer should be span-based or yes/no-based) and for a yes/no answer prediction:
\begin{equation*}
\begin{split}
\mathcal{L}_{type} = -\log\Bigg(\frac{\sum_{j\in P^Q} e^{s_{j}^{type}}}{\sum_{j\in P^Q} (e^{s_{j}^{binary}} + e^{s_{j}^{span}})}\Bigg)
\end{split}
\end{equation*}
\begin{equation*}
\begin{split}
\mathcal{L}_{yes/no} = -\log\Bigg(\frac{\sum_{j\in P^Q} e^{s_{j}^{yes/no}}}{\sum_{j\in P^Q} (e^{s_{j}^{yes}} + e^{s_{j}^{no}})}\Bigg),
\end{split}
\end{equation*}
where $P^Q$ is the set of paragraphs paired with the same question $Q$, and $e^{s_{j}^{binary}}, e^{s_{j}^{span}}$ and $e^{s_{j}^{type}}$ are the likelihood scores of the $j$-th question-paragraph pair being a binary yes/no-based type, a span-based type, and its true type, respectively. $e^{s_{j}^{yes}}, e^{s_{j}^{no}}$ and $e^{s_{j}^{yes/no}}$ are the likelihood scores of the $j$-th question-paragraph pair having the answer `yes', the answer `no', and its true answer, respectively. For span-based questions, $\mathcal{L}_{yes/no}$ is defined to be zero, and vice-versa. \par
For the supporting fact prediction, we use a binary cross-entropy loss on each sentence, $\mathcal{L}_{sp}$. The final loss function is the sum of these four objectives, $$\mathcal{L}_{hotpot} = \mathcal{L}_{span} + \mathcal{L}_{type} + \mathcal{L}_{yes/no} + \mathcal{L}_{sp}$$
During inference, the supporting facts prediction is taken only from the paragraph from which the answer is predicted.

\paragraph{Metrics}
Three sets of metrics were proposed by \citet{Yang0ZBCSM18hotpot} to evaluate performance on the HotpotQA dataset. The first set of metrics focuses on evaluating the answer span. For this purpose the exact match (EM) and \fone metrics are used, as suggested by \citet{RajpurkarZLL16squad}. The second set of metrics focuses on the explainability of the models, by evaluating the supporting facts directly using the EM and \fone metrics on the set of supporting fact sentences. The final set of metrics combines the evaluation of answer spans and supporting facts as follows. For each example, given its precision and recall on the answer span $(P^{\text{(ans)}}, R^{\text{(ans)}})$ and the supporting facts $(P^{\text{(sup)}}, R^{\text{(sup)}})$, respectively, the joint \fone is calculated as
$$P^{\text{(joint)}} = P^{\text{(ans)}}P^{\text{\text{(sup)}}}, R^{\text{(joint)}} = R^{\text{(ans)}}R^{\text{(sup)}},$$
$$\text{Joint \fone} = \frac{2P^{\text{(joint)}}R^{\text{(joint)}}}{P^{\text{(joint)}} + R^{\text{(joint)}}}.$$
The joint EM is 1 only if both tasks achieve an exact match and otherwise 0. Intuitively, these metrics penalize systems that perform poorly on either task. All metrics are evaluated example-by-example, and then averaged over examples in the evaluation set.

\section{Implementation Details}
\label{sec:implementation}
We use the Stanford CoreNLP toolkit \citep{ManningSBFBM14corenlp} for tokenization.
We implement all our models using TensorFlow. 
\paragraph{Architecture Details}
For the word-level embeddings, we use the GloVe 300-dimensional embeddings pretrained on the 840B Common Crawl corpus \citep{PenningtonSM14glove}. For the character-level embeddings, we use 20-dimensional character embeddings, and use a 1-dimensional CNN with 100 filters of size 5, with a dropout \citep{SrivastavaHKSS14dropout} rate of 0.2. \par
For the encoder, we also concatenate ELMo \citep{Peters2018elmo} embeddings with a dropout rate of 0.5 and the token representations from the output of embedding layer to form the final token representations, before processing them through the first bidirectional GRU. We use the ELMo weights pretrained on the 5.5B dataset.\footnote{Available at https://allennlp.org/elmo} To speed up computations, we cache the context independent token representations of all tokens that appear at least once in the titles of the HotpotQA Wikipedia version, or appear at least five times in the entire Wikipedia version. Words not in this vocabulary are given a fixed OOV vector. We use a learned weighted average of all three ELMo layers. Variational dropout \citep{variationalDropout}, where the same dropout mask is applied at each time step, is applied on the inputs of all recurrent layers with a dropout rate of 0.2.
We set the encoding size to be $d = 1024$. \par
For the paragraph reader used for HotpotQA, we use a state size of 150 for the bidirectional GRUs. The size of the hidden layer in the MLP used for supporting fact prediction is set to 150 as well. Here again variational dropout with a dropout rate of 0.2 is applied on the inputs of all recurrent layers and attention mechanisms. The reader used for SQuAD is the shared-norm model trained on the SQuAD dataset by \citet{GardnerC18snorm}.\footnote{Available at https://github.com/allenai/document-qa}

\paragraph{Training Details}
We train all our models using the Adadelta optimizer \citep{adadelta} with a learning rate of 1.0 and $\rho = 0.95$. \par
\textbf{SQuAD-Open:} The training data is gathered as follows.
For each question in the original SQuAD dataset, the original paragraph given as the question's context is considered as the single relevant (positive) paragraph. We gather $\sim$12 irrelevant (negative) paragraphs for each question in the following manner:
\begin{itemize}
    \item The three paragraphs with the highest TF-IDF similarity to the question in the same SQuAD document as the relevant paragraph (excluding the relevant paragraph). The same method is applied to retrieve the three paragraphs most similar to the relevant paragraph.
    \item The two paragraphs with the highest TF-IDF similarity to the question from the set of all first paragraphs in the entire Wikipedia (excluding the relevant paragraph's article). The same method is applied to retrieve the two paragraphs most similar to the relevant paragraph.
    \item Two randomly sampled paragraphs from the entire Wikipedia.
\end{itemize}
Questions that contain only stop-words are dropped, as they are most likely too dependent on the original context and not suitable for open-domain.
In each epoch, a question appears as a training sample four times; once with the relevant paragraph, and three times with randomly sampled irrelevant paragraphs. \par
We train with a batch size of 45, and do not use the ranking loss by setting $\lambda = 0$ in Equation (\ref{eq:sum_obj}). We limit the length of the paragraphs to 600 tokens. \par
\textbf{HotpotQA:} The paragraphs used for training the encoder are the gold and distractor paragraphs supplied in the original HotpotQA training set. As mentioned in Section \ref{sec:encoder}, each training sample consists of a question and two paragraphs, $(Q, P^1, P^2)$, where $P^1$ corresponds to a paragraph retrieved in the first iteration, and $P^2$ corresponds to a paragraph retrieved in the second iteration. For each question, we create the following sample types:
\begin{enumerate}
    \item Gold: The two paragraphs are the two gold paragraphs of the question. Both $P^1$ and $P^2$ are considered positive.
    \item First gold, second distractor: $P^1$ is one of the gold paragraphs and considered positive, while $P^2$ can be a random paragraph from the training set, the same as $P^1$, or one of the distractors, with probabilities 0.05, 0.1 and 0.85, respectively.
    $P^2$ is considered negative.
    \item First distractor, second gold: $P^1$ is either one of the distractors or a random paragraph from the training set, with probabilities 0.9 and 0.1, respectively. $P^2$ is one of the gold paragraphs. Both $P^1$ and $P^2$ are considered negative.
    \item All distractors: Both $P^1$ and $P^2$ are sampled from the question's distractors, and are considered negative.
    \item Gold from another question: A gold paragraph pair taken from another question; both paragraphs are considered negative.
\end{enumerate}
The use of the sample types from the above list motivation is motivated as follows. Sample type 1 is the only one that contains purely positive examples and hence is mandatory. Sample type 2 is necessary to allow the model to learn a valuable reformulation, which does not give a relevant score based solely on the first paragraph. Sample type 3 is complementary to type 2; it allows the model to learn that a paragraph pair is irrelevant if the first paragraph is irrelevant, regardless of the second. Sample type 3 is used for random negative sampling, which is the most common case of all. Sample type 4 is used to guarantee the model does not determine relevancy solely based on the paragraph pair, but also based on the question. \par
In each training batch, we include three samples for each question in the batch: a single gold sample (type 1), and two samples from the other four types, with sample probabilities of 0.35, 0.35, 0.25 and 0.05, respectively. 
\par 
We use a batch size of 75 (25 unique questions). We set the margin to be $\gamma = 1$ in Equation (\ref{eq:rank_obj}) and $\lambda = 1$ in Equation (\ref{eq:sum_obj}), for both prediction iterations. We limit the length of the paragraphs to 600 tokens. \par
\textbf{HotpotQA Reader:} The reader receives a question and a concatenation of a paragraph pair as input. Each training batch consists of three samples with three different paragraph pairs for each question: a single gold pair, which is the two gold paragraphs of the question, and two randomly sampled paragraph pairs from the set of the distractors and one of the gold paragraphs of the question. We label the correct answer spans to be every text span that has an exact match with the ground truth answer, even in the distractor paragraphs. We use a batch size of 75 (25 unique questions), and limit the length of the paragraphs (before concatenation) to 600 tokens.

\end{document}